\documentclass[pamm,a4paper,fleqn]{w-art}
\usepackage{times,cite,w-thm}
\usepackage[T1]{fontenc}
\usepackage[utf8]{inputenc}
\usepackage{graphicx}
\usepackage{amsmath}
\usepackage{amsfonts}
\usepackage{subcaption}
\usepackage{wrapfig}
\usepackage{mathtools}

\begin{document}

\TitleLanguage[EN]
\title[The short title]{Applied Bayesian Structural Health Monitoring: inclinometer data anomaly detection and forecasting}

\author{\firstname{David K. E.} \lastname{Green}\inst{1}%
\footnote{Corresponding author: e-mail \ElectronicMail{david.green@dywidag.com}} 
\address[\inst{1}]{\CountryCode[UK]DYWIDAG Ltd UK, Datum House, The Pavilions, Bury BL9 7NY, UK}}
\author{\firstname{Adam} \lastname{Jaspan}\inst{1}}
\AbstractLanguage[EN]
\begin{abstract}
Inclinometer probes are devices that can be used to measure deformations within earthwork slopes. This paper demonstrates a novel application of Bayesian techniques to real-world inclinometer data, providing both anomaly detection and forecasting. Specifically, this paper details an analysis of data collected from across the entire UK rail network. 

Inclinometers are a standard tool of geotechnical site monitoring. Data from these instruments is often used in risk analysis and decision-making. However, discerning anomalous data points and forecasting future behaviour from inclinometer data requires significant `engineering judgement' (subjective appraisal). This is because the observational data is derived from complex physical phenomena and contains complex spatio-temporal correlations. Additionally, the practical demands of data collected from remote sites over several years tends to introduce systematic errors. These issues make the interpretation of inclinometer data challenging.

Practitioners have effectively two goals when processing monitoring data. The first is to identify any anomalous or dangerous movements, and the second is to predict potential future adverse scenarios by forecasting. In this paper we apply Uncertainty Quantification (UQ) techniques by implementing a Bayesian approach to anomaly detection and forecasting for inclinometer data. Subsequently, both costs and risks may be minimised by quantifying and evaluating the appropriate uncertainties. This framework may then act as an enabler for enhanced decision making and risk analysis.

We show that inclinometer data can be described by a latent autocorrelated Markov process derived from measurements. This can be used as the transition model of a non-linear Bayesian filter. This allows for the prediction of system states. This learnt latent model also allows for the detection of anomalies: observations that are far from their expected value may be considered to have `high surprisal', that is they have a high information content relative to the model encoding represented by the learnt latent model. 

We successfully apply the forecasting and anomaly detection techniques to a large real-world data set in a computationally efficient manner. Although this paper studies inclinometers in particular, the techniques are broadly applicable to all areas of engineering UQ and Structural Health Monitoring (SHM).  
\end{abstract}
\maketitle

\section{Introduction}

\subsection{Overview}

Globally, civil public infrastructure is ageing rapidly. Climate change and population growth are expected to accelerate the rate of deterioration and therefore increase the maintenance costs of such infrastructure. Maintenance of degraded or failed sites is typically far more expensive than pre-emptive repair and stabilisation. Traditionally, SHM - the use of sensor equipment to observe civil infrastructure - has been utilised to provide insights into the performance of structures \cite{soga2016infrastructure}. However, traditional approaches to SHM primarily report the readings from sensors and do not apply any interpretation of the data. Advances in the field have been largely restricted to improved data collection and presentation techniques (such as remote sensors, cloud data storage and web-based data visualisations \cite{II}). Interpretation of the data is left to practitioners. When applied to the vast amounts of data generated by SHM, this may lead to human error, and places additional strain on maintenance budgets.

In this paper we demonstrate that statistical and probabilistic UQ approaches can be used to understand and interpret inclinometer data in an automated fashion, easing the burden on practitioners and releasing resources for infrastructure maintenance. Probabilistic latent state models can be used to extract a signal from noisy data in a consistent, repeatable way without reliance on subjective engineering judgement. 

A particular advantage of employing a probabilistic approach is that forecasts of future system states can be made using Recursive Bayesian Filtering \cite{murphy2012machine, sarkka2013bayesian, russell2010artificial}. Forecasting allows for early warning alerts to be provided. For example, a forecast might predict that deformations of a structure may exceed an acceptable limit at some time, perhaps several months, in the future. Such early warning alerts would allow for structural remediation to occur well in advance of any structural failures. Early maintenance of a degraded structure is far less expensive than repair of failed infrastructure.

\begin{figure}[hb!]
  \centering
  \begin{subfigure}{.5\textwidth}
    \centering
    \includegraphics[height=5cm]{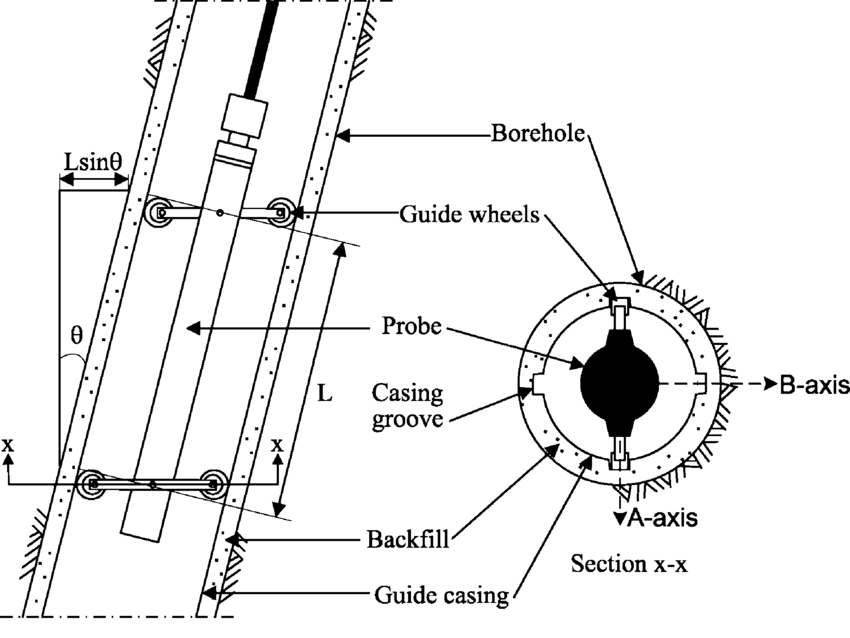}
    \caption{Inclinometer configuration.}
    \label{fig:inclo_sub1}
  \end{subfigure}%
  \centering
  \begin{subfigure}{.5\textwidth}
    \centering
    \includegraphics[height=5cm]{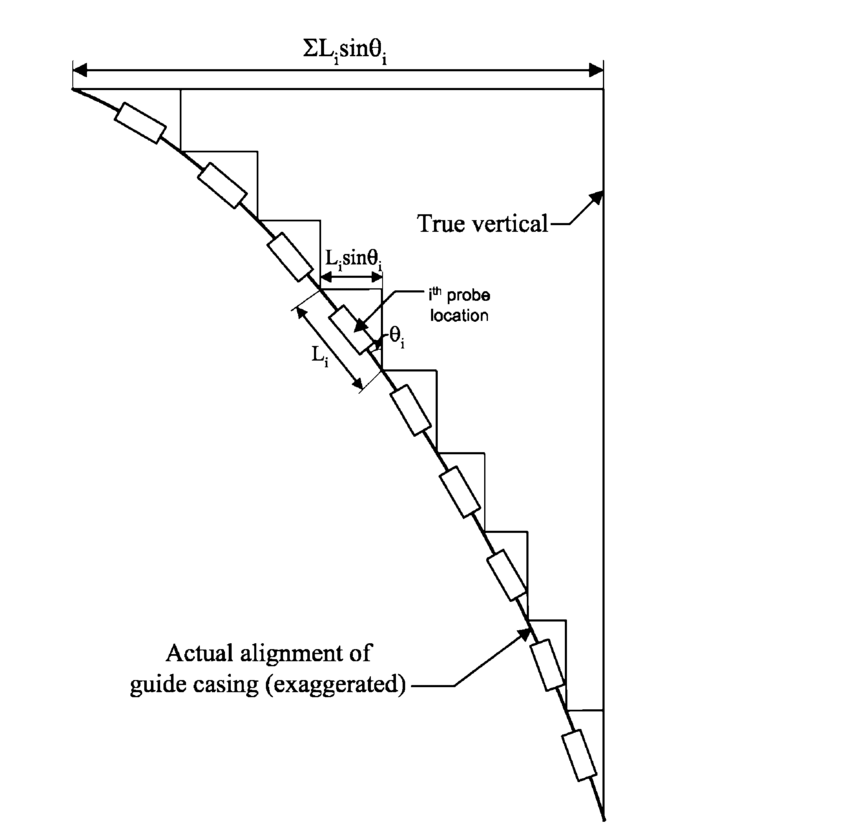}
    \caption{Inclinometer deflection measurement.}
    \label{fig:inclo_sub2}
  \end{subfigure}%
  \caption{Inclinometer illustrations, from \cite{stark2008slope}}
  \label{fig:inclinometer}
\end{figure}

An inclinometer is a measurement device that records the horizontal deflection of the subsurface \cite{machan2008use}. Measurements of subsurface deflection are collected at fixed depth increments down a pre-drilled borehole. Inclinometers are commonly used to measure the displacements and stability of embankments along road and rail corridors. Fig. \ref{fig:inclinometer} provides a depiction of an inclinometer device. Inclinometers record deformations relative to a set of orthogonal basis vectors, the so-called $A$ and $B$ axes respectively. The $A$ axis is oriented to increase along the steepest slope of the ground surface. The $B$ axis is set orthogonal to the $A$ axis. Data from both manual and so-called `in-place' inclinometers are analysed in this paper. In-place inclinometers record their data automatically at more regular intervals but are more expensive to install and maintain than manual inclinometers and are typically used only for critical sites. The installation and type of the inclinometer used affects the data quality. By applying UQ techniques, we can automatically extract the relevant signals from the inclinometer measurements regardless of the device type.

Kalman Filtering is used to model the inclinometer data as a Hidden Markov Model \cite{russell2010artificial}. A fixed, kinematically derived transition matrix is used. An observation model that accounts for the manufacturer-provided instrument accuracy is included in the model. The Kalman Filter model is enhanced with an RTS smoother \cite{rauch1965maximum, sarkka2008unscented} to infer both latent states and the a priori unknown process covariance. We introduce a simple, computationally inexpensive approach to dealing with data that has been recorded at irregular time intervals. Real-world data frequently features outliers. We detect and exclude these outliers using the probabilistic latent state formulation presented.

To verify our analysis, we test on real-world data using a hold-out validation set. Our automated process was able to analyse over 10000 manual inclinometer readings from 20 boreholes over 4 sites and over 12 million in-place inclinometer readings from 10 boreholes over 5 sites. We demonstrate that our Kalman Filter approach can effectively model the latent state and produce valid data forecasts. We demonstrate that the Bayesian data analysis presented in this paper is a significant improvement over existing techniques for inclinometer data processing and presentation.

\section{Filter representation}

\subsection{Latent state representation}

As an inclinometer can be modelled as a classical physical system, a kinematic transition model is used. At each depth in the borehole, $i \in [1,n]$, the discrete time evolution of the system (excluding noise) is modelled by
\begin{align}
q^i_k = q^i_k + p^i_k \Delta t + w_k
\end{align}
where $w_k$ is the process noise, $\Delta t \in \mathbb{R} > 0$ is the time, $t$, difference between each $q_k = q(t)$ and $p_k = \frac{d}{dt}q_k$. Note that $p, q \in \mathbb{R}^2$ representing the $A$ and $B$ axes of the inclinometer, then $q^i_k = (q^i_{k,A}, q^i_{k,B})$ and $p^i_k = (p^i_{k,A}, p^i_{k,B})$.

The state space is then taken to be:
\begin{align}
  \label{eqn:statespacemodel}
  x_k = \begin{bmatrix}
    q^1_k, \cdots, q^n_k, 
    p^1_k, \cdots, p^n_k  \end{bmatrix}^T = \begin{bmatrix}
      q^1_{k,A}, q^1_{k,B}, \cdots, q^n_{k,A}, q^n_{k,B},
      p^1_{k,A}, p^1_{k,B}, \cdots, p^n_{k,A}, p^n_{k,B} \end{bmatrix}^T \in \mathbb{R}^{4n}.
\end{align}

\subsection{Kalman Filter state representation}

To model and forecast the inclinometer data, a discrete-time Kalman Filter-type Hidden Markov Model \cite{russell2010artificial} is used:
\begin{align}
  \label{eqn:futureStateEvolution}
  x_k = F_k x_{k-1} + w_k
\end{align}
where:
\begin{itemize}
  \item $F_k \in \mathbb{R}^{4n \times 4n}$ is the state transition model.
  \item $w_k \sim \mathcal{N}(0, Q_k)$ is the process noise with so-called `process covariance' $Q_k$.
\end{itemize}
Each latent state, $x_k$, is normally distributed: $x_k \sim \mathcal{N}\left(\mu_k, \Sigma_k\right)$. Note that while a typical Kalman Filter model includes control inputs, there are no control inputs present or applied for the inclinometer system. 

Entries of the transition matrix are given by
\begin{align}
  F_{k,ij} =      
  \begin{cases}
    i < 2n & \begin{cases} j < 2n & \delta(i,j) \\ \text{else} & \delta(i,j-2n) \Delta t \\ \end{cases} \\\\
    \text{else} & \begin{cases} j < 2n & 0 \\ \text{else} & \delta(i,j) \\ \end{cases} \\
  \end{cases}
\end{align}
where, for $i, j \in \mathbb{Z}$, $\delta(i,j) = 1$ if $i = j$ and $0$ otherwise and the order of the state space model degrees of freedom is as given in eqn. \ref{eqn:statespacemodel}. For an example of a kinematic transition matrix, see \cite{chen2021time}.

For notational convenience, assume that forecasted latent states (that is, states estimated at times into the future beyond the current observational data using eqn. \ref{eqn:futureStateEvolution}) are denoted $x_f$. Let the transition matrix for forecasting states be given by $F_f$ with process covariance $Q_f$.

\subsection{Kalman Filter observation representation}

At any time $k$ an observation, $z_k$, is given by $z_k = H_k x_k + v_k$ where $H_k \in \mathbb{R}^{2n\times 4n}$ is a fixed observation model with entries $H_{k,ij} = \delta(i,j)$. The term $v_k$ represents the observation noise. Typically, in a Kalman Filter, $v_k \sim \mathcal{N}(0, R_k)$ is assumed. As velocities are not measured by an inclinometer, a degenerate form of $R_k$ is used where the variance (and covariance) of any velocity degrees of freedom is assumed to be $0$. For the positional degrees of freedom, $q$, the scale of $R_k$ is set to $\epsilon_m d$ where $\epsilon_m \in \mathbb{R}$ is the manufacturer-provided error as a function of depth of the inclinometer instrument and $d \in \mathbb{R}$ is the depth into the borehole. That is $R_{k,ij} = \delta(i,j) (\epsilon_m d)^2$ for $R_{k} \in \mathbb{R}^{2n \times 2n}$. Note that as $R_k$ is time independent it can be written $R$.

\subsection{Expectation Maximisation and the RTS smoother}

Expectation Maximisation is a probabilistic technique to find a maximum likelihood estimate of latent parameters from observations \cite{murphy2012machine}. For the Kalman Filtering problem in this paper, the latent parameters are the state estimates at each time and the process covariance matrix, $Q_k$. Estimation of the latent state parameters at time states earlier than the most recent observation is referred to as `smoothing' \cite{murphy2012machine}. The RTS Smoother \cite{rauch1965maximum} can be considered an extension of the classic Baum-Welch algorithm \cite{russell2010artificial} to a specific case of continuous-valued latent state Hidden Markov Model. 

The computational complexity of an RTS smoother is at least $\mathcal{O}(m n^3)$ where $m$ is the number of data points. As the amount of data grows over time, the run time of the smoother will increase. To prevent this, we use only a finite number of previous data points when applying the smoother. 

Of note is that the standard form of RTS smoother we use does not account for constraints on the form of $Q_k$ and assumes a fixed $\Delta t$. The significance of this will be discussed in more detail throughout this paper.

\section{Processing observational data with irregular sampling frequencies}

\subsection{Observation data sampling frequency irregularities}

The real-world data analysed in this paper is not sampled with perfect regularity. As such, it is necessary to either alter the standard, discrete time formulation of a Kalman Filter or to modify the data such that $\Delta t$ is constant. Additionally, RTS smoothing requires a fixed $\Delta t$. 

A simple approach to regularise the data spacing would be to interpolate the readings at their true sample times to a new, regularly spaced $\Delta t$. This would function as a low-pass filter and reduce the variance of the estimated latent state. This would be non-conservative (in the structural stability sense, as deformations would be underestimated) and therefore undesirable. To reduce (but not eliminate) this sort of error we instead use a form of data upscaling. A time increment is chosen per borehole (data measurement set) and is typically approximately $\min_i \| t_{i+1} - t_{i}\|^2$. The selection of $\Delta t$ is a trade-off between accuracy and processing time. 

The time of each observation is then mapped on to the closest time increment, $j$, such that $(t_i - j\Delta t)$ is minimised. By upscaling the data, we avoid cutting the peaks off the observations and therefore avoid underestimating the variance in the latent state.

For the actual data analysed the lack of regularity in the sampling rate is small and as such the error induced by observation time remapping is assumed to be negligible compared to the other sources of noise present in the system. In \cite{chen2021time} a method for fully time-variable Kalman Filtering is presented. We leave an implementation of such a method for future work as the computational complexity of such a method is expected to be significantly higher than the RTS method.

\subsection{Dependence of process covariance on $\Delta t$}
\label{ssec:processcovardt}

For forecasting, using discrete time evolution operations, $\Delta t$ is also required. However, it is undesirable (for the particular data we have) that the $\Delta t$ used for forecasting is as large as the time difference between observations as the value of $\Delta t$ alters the rate of growth of numerical errors when computing the forward-time evolution of the system While $\Delta t$ is easily modified in the transition matrix, $F_k$, the $Q_k$ learnt by the RTS smoother is dependent on $\Delta t$ in a non-trivial way. We demonstrate a technique to infer a kinematically valid $Q_f$ for forecasts after learning $Q_k$ with an RTS smoother.

The estimation of process covariance from observations is a significant challenge and an open area of research. For a kinematic latent time evolution model the process covariance can be shown to depend on $\Delta t$ \cite{chen2021time,sarkka2013bayesian} as follows:
\begin{align}
  Q_k &= \sigma_\alpha \Lambda \\
  \Lambda_{ij} &= 
  \begin{cases}
    i < n & \begin{cases}
      j < n & \frac{\Delta t^3}{3}\delta(i,j) \\
      \text{else} & \frac{\Delta t^2}{2}\delta(i,j-n) \\
    \end{cases} \\
    \\
    \text{else} &  
    \begin{cases}
        j < n & \frac{\Delta t^2}{2}\delta(i-n,j) \\
        \text{else} & \Delta t\delta(i,j) \\
      \end{cases} \\
  \end{cases}
\end{align}
where $\sigma_\alpha \in \mathbb{R}$ represents the intensity of the process covariance white noise. This representation of $Q$ in terms of $\Lambda$ is derived by using a matrix exponential representation of the continuous time evolution of the latent model and then approximating this exponential via Van Loan's method (a form of Taylor series expansion of the matrix exponential) \cite{horn2012matrix}.

As the RTS smoother learns the covariance matrix, $Q_k$, it is necessary to fix the $\Delta t$ to ensure that the time dependence for $Q_k$ is fixed. For forecasting, it is desirable to modify $\Delta t$. For a given $\Delta t$, $\Lambda$ can be computed but $\sigma_\alpha$ must be solved for. We find $\sigma_\alpha$ by minimising the matrix 2-norm
\begin{align}
  \| Q_k - \sigma_\alpha \Lambda(\Delta t)\|_2 = \left( \sum_i^{4n} \sum_j^{4n} Q_{k,ij} - \sigma_\alpha \Lambda_{ij} \right)^2.
\end{align}
Using a simple quadratic least squares approach, setting $
  \frac{\partial}{\partial \sigma_\alpha}\| Q_k - \sigma_\alpha \Lambda(\Delta t)\|^2 = 0$ and solving for $\sigma_\alpha$ yields:
\begin{align}
  \sigma_\alpha = \frac{ \sum_i^{4n} \sum_j^{4n}Q_{k,ij}\Lambda_{ij}}{\sum_i^{4n} \sum_j^{4n} \Lambda_{ij}^2}.
\end{align}
With the approximated $\sigma_\alpha$, the kinematic $Q_f = \sigma_\alpha \Lambda$ can be used for forecasting with any valid $\Delta t$, independent of the timestep used for smoothing.

\section{Anomaly detection}

Real-world data collection may contain significant errors which are not representative of the true state of the system. These errors may be caused by, for example, operator error or equipment failure. Discarding these outliers can improve the accuracy of latent state estimates. A latent probabilistic representation of measurement data has the advantage that the probability to observe a data point can be computed explicitly. We define an anomaly to be any point with a low probability to be observed given the previous latent state and the forecast model, $P(z_{k+1}) < \gamma \in [0, 1.0] \in \mathbb{R}$. We demonstrate a validation gating technique \cite{spehn1990noise, labbe2014kalman} with relatively (\cite{sun2020new}) low computational cost.

To compute the probability of an observation, $z_{k+1}$, the latent state estimate at $k$ is forecast forward to time $k+1$ and then mapped on to the observation space:
\begin{align}
  P(z_k) &= \int_x P(z_k|x_k)P(x_k) \\
  &= \mathcal{N}\left(H\mu_k, H\Sigma_k H^T + R \right)
\end{align}

In practice, the Mahalanobis distance \cite{stats:2008} is used in place of $P(z_k)$ as this is proportional to $P(z_k)$:

\begin{align}
  d_m(X, \mu, \Sigma) = \sqrt{\left( X - \mu \right)^T \Sigma^{-1} \left(X - \mu\right)}
\end{align}
where $\mu$ and $\Sigma$ are, respectively, the mean and covariance matrix of a multivariate normal. An anomaly is considered to be a point $X$ such that $d(x_k, H\mu_k, H\Sigma_kH^T + R) > \gamma$ for $x_k \sim \mathcal{N}\left(\mu_k, \Sigma_k\right)$.

\section{Testing and implementation}
\label{sec:testingandimplementation}

\subsection{Validation metric}

The methodology described in the previous sections was validated against a large set of real data. A data set consisting of over 12 million measurements from 31 boreholes taken over the last decade have been tested. It was anticipated that the manual inclinometer readings would feature more noise and outliers compared to the automated in-place inclinometers.

The forecasting technique was tested by computing the KL divergence, denoted $D_{KL}( \cdot \| \cdot)$ \cite{murphy2012machine}, between a forecast and a validation data set for each borehole. For each borehole a set of observations from times $k$ to $k+m$ were selected: $z_{k:k+m} \coloneqq 
(z_k, z_{k+1}, \dots, z_{k+m})$. The time range, $k$ to $k+m$ was chosen so that no anomalies were present in the validation set. The observations were split into a validation set of size $n$, $z_{k+m-n:k+m}$ and a training set: $z_{k:k+m-n}$. The forecast time range, $m$, was selected to allow for a one month prediction period. 

The quality of the forecast estimates was assessed by first smoothing to compute the latent state model for the training set, $x_{k:k-m+n}$. The latent model for the full data set (training and validation data) was also computed by smoothing: $x^v_{k:k+m}$. A forecast, $\hat{x}_{k:k-m+n}$, was computed using the latent model inferred from the training data. The forecast computed with the training model was compared to the estimate of the validation set latent state:
\begin{align}
  \label{eqn:verificatonmetric}
  F[z_{k:k+m}] = \frac{1}{m}\frac{1}{n}\sum^{k+m}_{i=k-m} \sum^n_{j=0} D_{KL}\left(\hat{x}_{i,j} \| x^v_{i,j} \right)
\end{align}
where $n$ denotes the number of measured depths and $x_{i,j}$ denotes the latent state at time $i$ at depth $j$.

\subsection{Observations}

\begin{figure}[b]
  \centering
    \includegraphics[height=5.5cm]{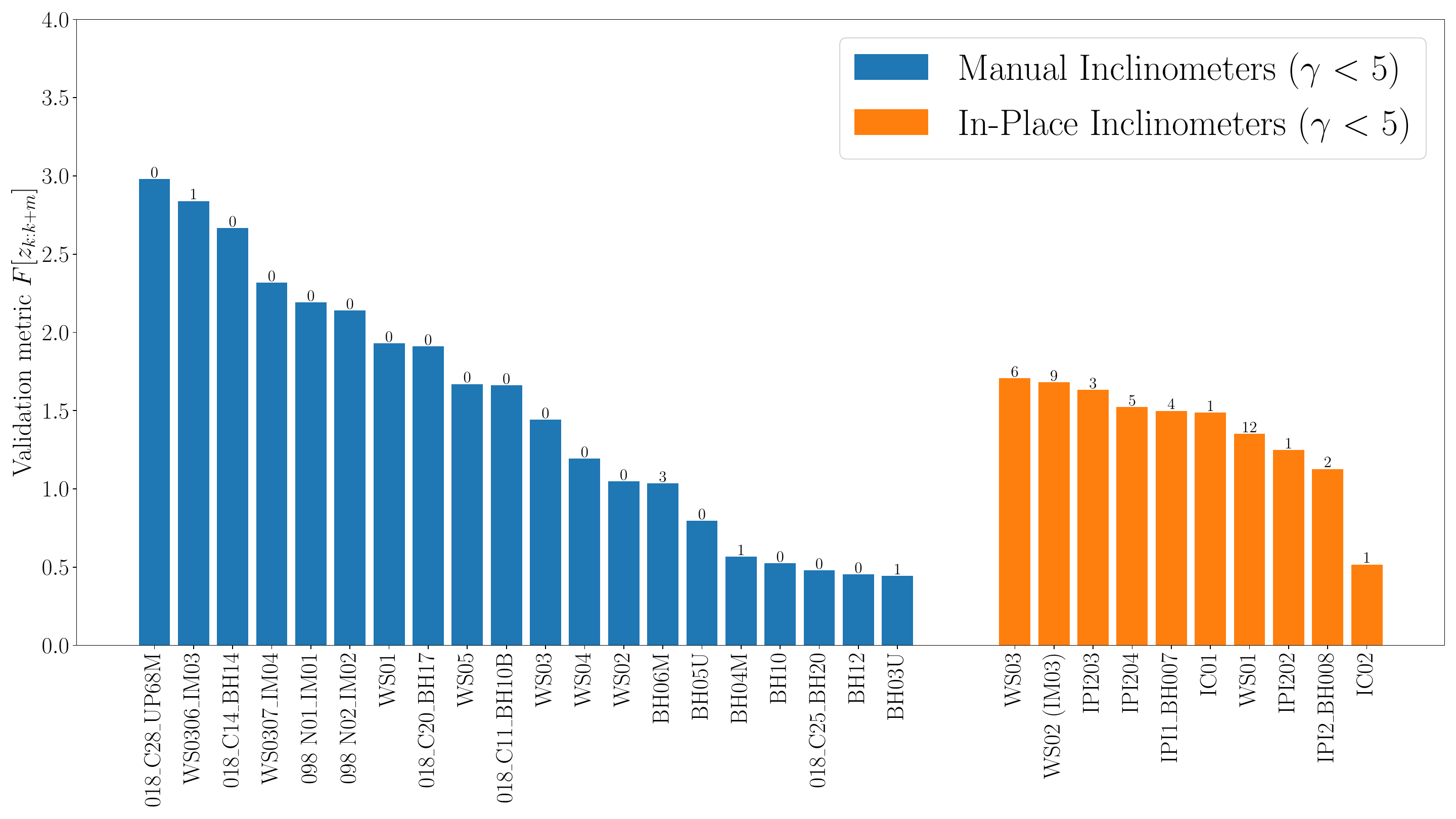}
  \caption{Summary of manual and in-place inclinometer verificaiton metric (eqn. \ref{eqn:verificatonmetric}), results for anomaly detection threshold $\gamma < 5$.  The name of each borehole data set is given under the corresponding bar. The number over each bar denotes the number of anomalies detected and removed in the training set.}
  \label{fig:kldivs}
\end{figure}

\begin{figure}[b]
  \centering
  \begin{subfigure}{.5\textwidth}
    \centering
    \includegraphics[height=.45\linewidth]{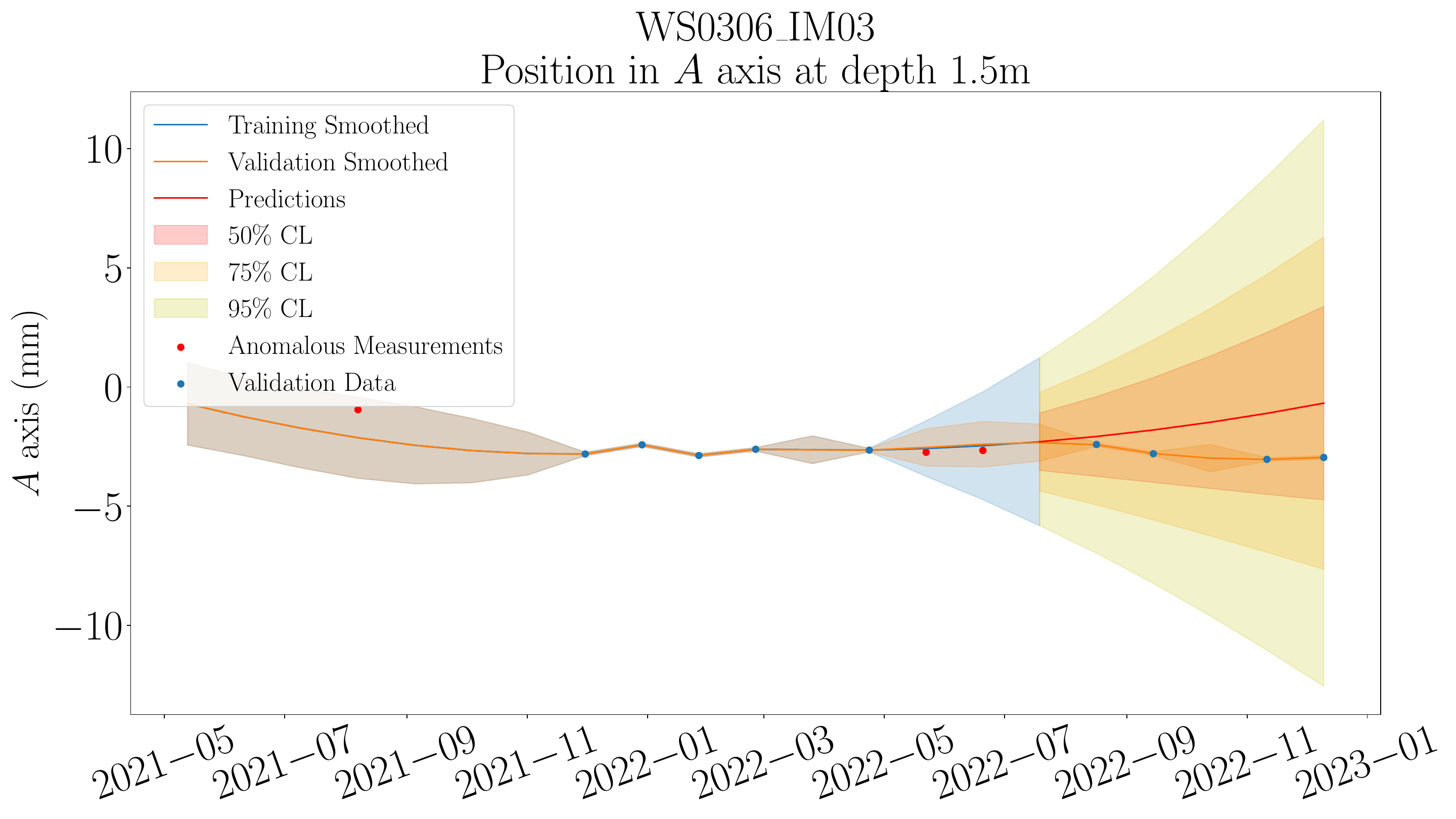}
    \caption{A axis time series plot.}
    \label{fig:sub1}
  \end{subfigure}%
  \begin{subfigure}{.5\textwidth}
    \centering
    \includegraphics[height=.45\linewidth]{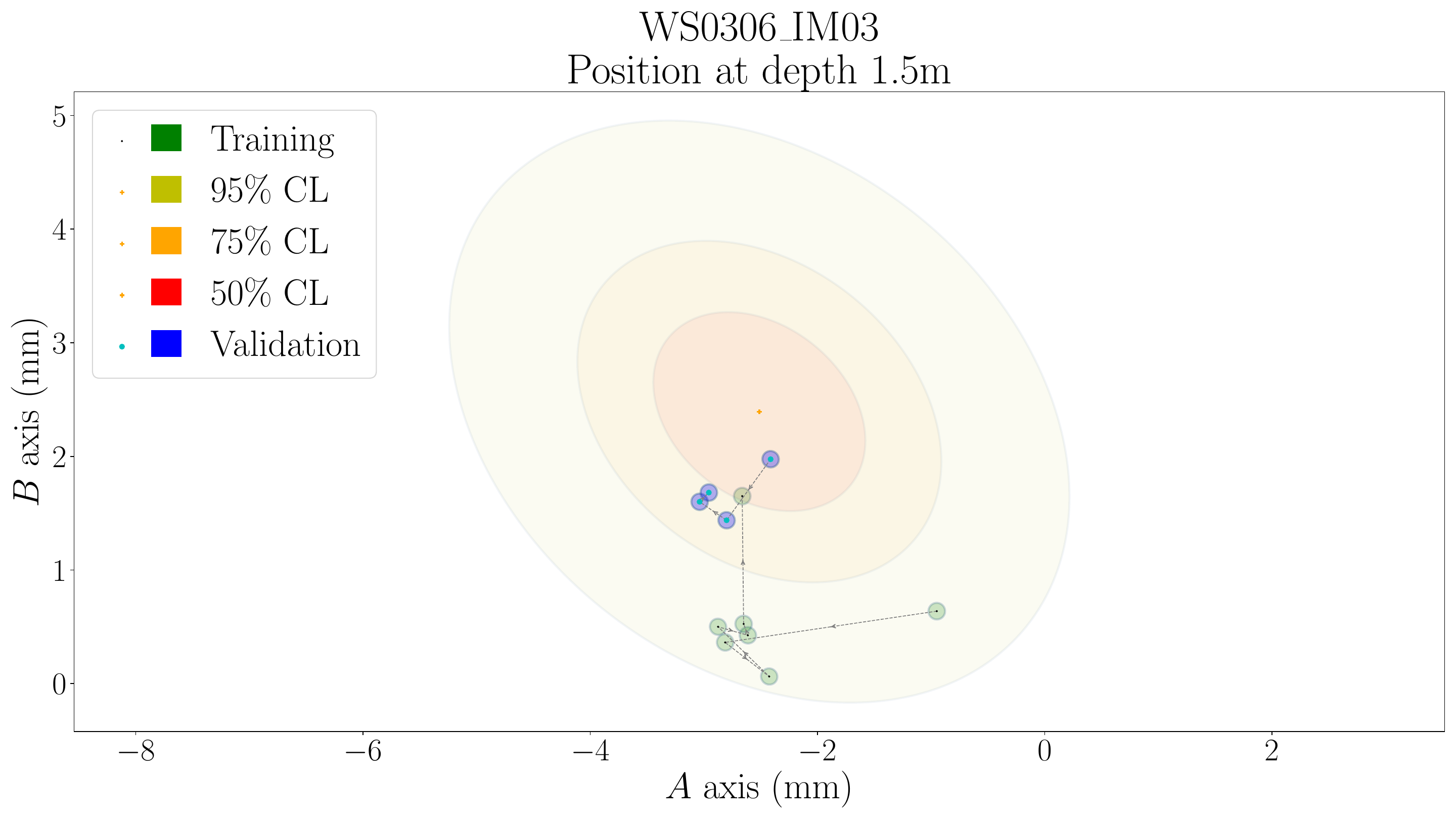}
    \caption{Combined A and B axis plot.}
    \label{fig:sub2}
  \end{subfigure}
  \caption{Example one month forecast result from borehole WS0306 IM03 at 1.5 m. The forecast uncertainty bands capture the variability in the validation data.}
  \label{fig:exampleborehole}
\end{figure}

A summary of the verification metric based on the KL divergence of training and test data computed across all sample boreholes is shown in Fig. \ref{fig:kldivs}. Higher values indicate a worse prediction. We note that the automated inclinometers are often able to produce better forecasts than manual inclinometers. This is as anticipated as the automated inclinometers record more data more accurately than the manual devices. A specific example forecast is shown in Fig. \ref{fig:exampleborehole}. In that example, the uncertainty bands can be seen to capture the variability in the validation data over a one month forecast period.

\begin{figure}[b]
  \centering
  \begin{subfigure}{.5\textwidth}
    \centering
    \includegraphics[height=.45\linewidth]{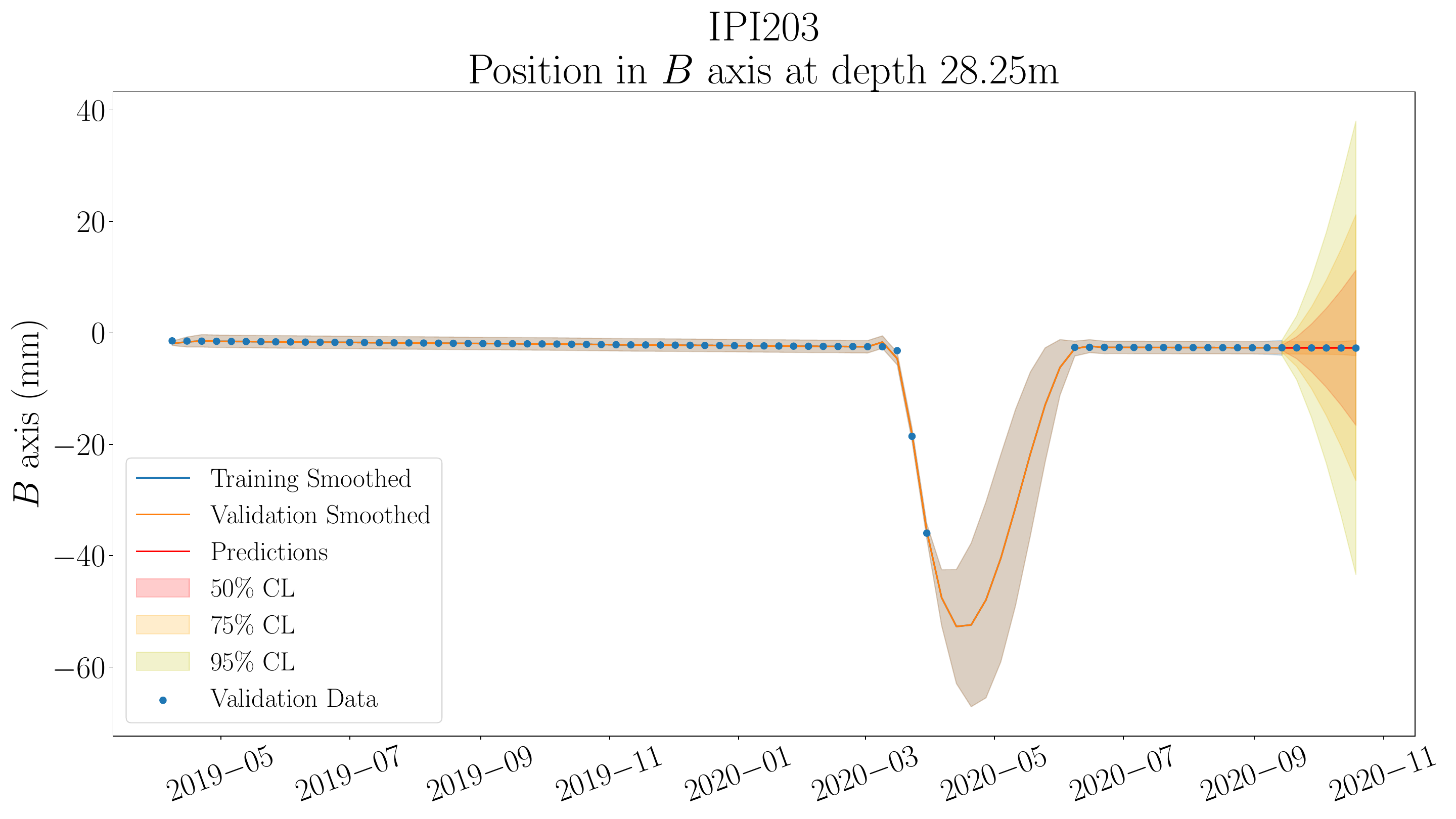}
    \caption{No anomaly detection.}
    \label{fig:sub1noanomaly}
  \end{subfigure}%
  \begin{subfigure}{.5\textwidth}
    \centering
    \includegraphics[height=.45\linewidth]{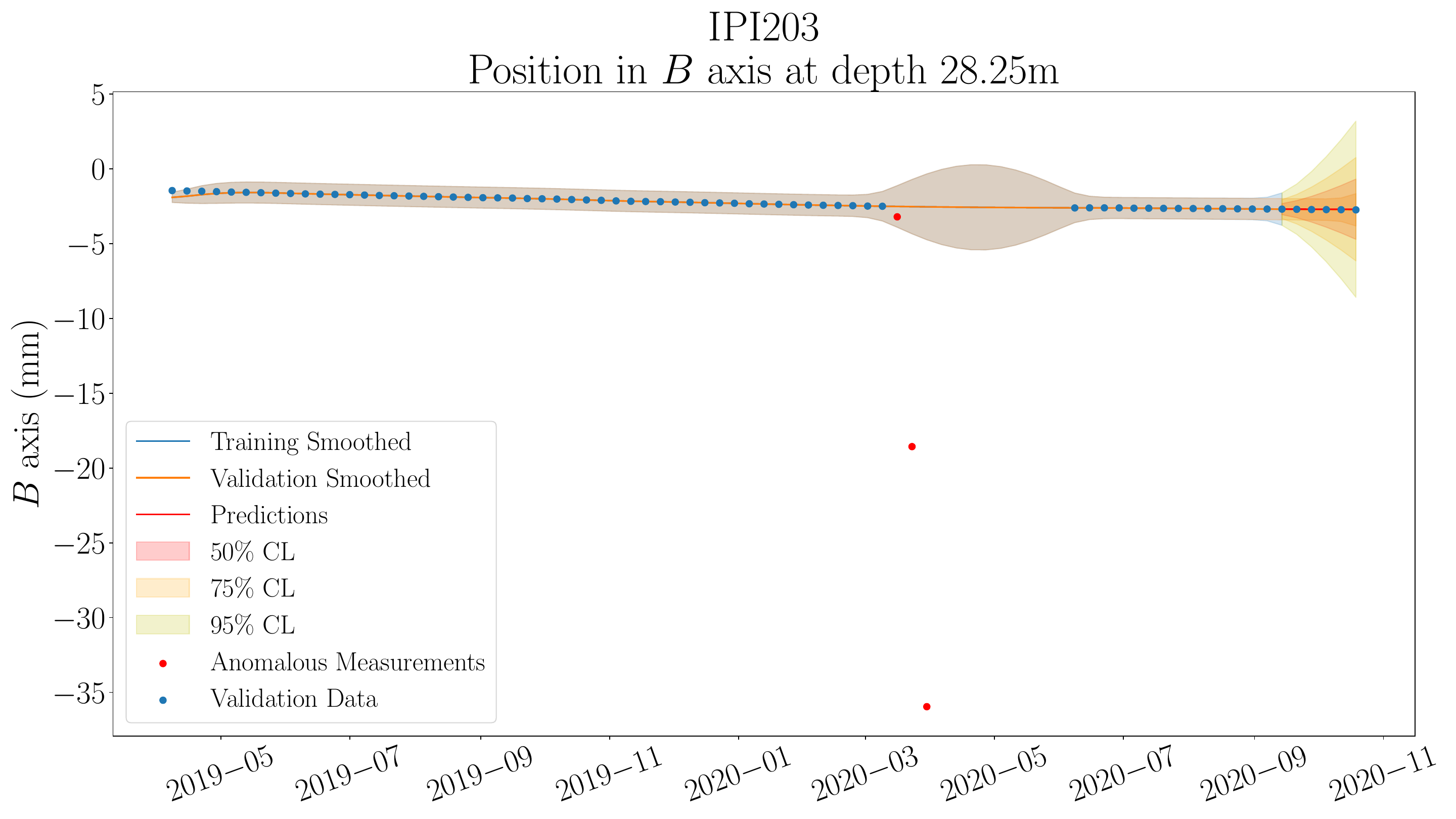}
    \caption{Anomaly detection $\gamma < 5$ enabled.}
    \label{fig:sub2yesanomaly}
  \end{subfigure}
  \caption{Example result from borehole IPI204 at 28.25 m depth demonstrating the utility of anomaly detection. The anomalous measurement in March 2020 was found to be the result of equipment failure.}
  \label{fig:anomalydetection}
\end{figure}

The utility of anomaly detection is demonstrated in Fig. \ref{fig:anomalydetection} which shows the $A$ axis' time series plot for inclinometer data filtering both with and without anomaly detection. The anomalous measurement in March 2020 is known to have been caused by equipment failure. The anomaly detection algorithm was able to detect and exclude invalid data points.

\section{Conclusions}

This paper presented a Bayesian approach to the analysis of real-world inclinometer data. Several extensions to the model presented are left for future work. For the inclinometer data analysed in this paper, the model was able to produce valid forecasts, up to one month into the future. Although not presented, the model is in fact capable of even longer range forecasts under the right conditions. The ability to forecast far into the future could be used to provide early warning of excessive deformations to asset owners, allowing for predictive and proactive maintenance scheduling.

The approach adopted is computationally efficient. In particular, a novel approach to modifying the process covariance of a data set learnt with an EM-based RTS smoother for forecasting was presented. Using this technique, the computational benefits of the EM smoother were retained while still allowing for a kinematically-correct process covariance to be used for forecasting. 

We believe that the Bayesian filtering approach to inclinometer data processing is a significant improvement over standard SHM data techniques as it allows for the automation of a substantial part of traditional data interpretation, allowing for large volumes of data to be analysed in a cost-effective manner.

\vspace{\baselineskip}

\bibliographystyle{plain} 
\bibliography{refs} 

\begin{thebibliography}{10}

\bibitem{stats:2008}
{\em The Concise Encyclopedia of Statistics}.
\newblock Springer New York, New York, NY, 2008.

\bibitem{chen2021time}
Zhaozhong Chen, Christoffer Heckman, Simon Julier, and Nisar Ahmed.
\newblock Time dependence in kalman filter tuning, 2021.

\bibitem{II}
DYWIDAG.
\newblock {I}nfrastructure {I}ntelligence.
\newblock \url{http://https://dywidag.com/monitoring}.

\bibitem{horn2012matrix}
Roger~A Horn and Charles~R Johnson.
\newblock {\em Matrix analysis}.
\newblock Cambridge university press, 2012.

\bibitem{labbe2014kalman}
Roger Labbe.
\newblock Kalman and bayesian filters in python.
\newblock {\em Chap}, 7(246):4, 2014.

\bibitem{machan2008use}
George Machan and Victoria~Gene Bennett.
\newblock Use of inclinometers for geotechnical instrumentation on
  transportation projects: State of the practice.
\newblock {\em Transportation Research Circular}, (E-C129), 2008.

\bibitem{murphy2012machine}
Kevin~P Murphy.
\newblock {\em Machine learning: a probabilistic perspective}.
\newblock MIT press, 2012.

\bibitem{rauch1965maximum}
Herbert~E Rauch, F~Tung, and Charlotte~T Striebel.
\newblock Maximum likelihood estimates of linear dynamic systems.
\newblock {\em AIAA journal}, 3(8):1445--1450, 1965.

\bibitem{russell2010artificial}
Stuart~J Russell.
\newblock {\em Artificial intelligence a modern approach}.
\newblock Pearson Education, Inc., 2010.

\bibitem{sarkka2008unscented}
Simo S{\"a}rkk{\"a}.
\newblock Unscented rauch--tung--striebel smoother.
\newblock {\em IEEE transactions on automatic control}, 53(3):845--849, 2008.

\bibitem{sarkka2013bayesian}
Simo S{\"a}rkk{\"a}.
\newblock {\em Bayesian filtering and smoothing}.
\newblock Number~3. Cambridge University Press, 2013.

\bibitem{soga2016infrastructure}
Kenichi Soga and Jennifer Schooling.
\newblock Infrastructure sensing.
\newblock {\em Interface focus}, 6(4):20160023, 2016.

\bibitem{spehn1990noise}
Stephen~L Spehn.
\newblock Noise adaptation and correlated maneuver gating of an extended kalman
  filter.
\newblock Technical report, Naval Postgraduate School Monterary CA, 1990.

\bibitem{stark2008slope}
Timothy~D Stark and Hangseok Choi.
\newblock Slope inclinometers for landslides.
\newblock {\em Landslides}, 5:339--350, 2008.

\bibitem{sun2020new}
Jinping Sun, Ziwei Wang, and Qing Li.
\newblock A new multiple hypothesis tracker using validation gate with motion
  direction constraint.
\newblock {\em Sensors}, 20(17):4816, 2020.

\end{thebibliography}

\end{document}